\documentclass[conference]{IEEEtran}
\IEEEoverridecommandlockouts
\usepackage{cite}
\usepackage{amsmath,amssymb,amsfonts}
\usepackage{algorithmic}
\usepackage{graphicx}
\usepackage{textcomp}
\usepackage{hyperref}
\usepackage{xcolor}
\usepackage{multirow}
\usepackage{svg}
\def\BibTeX{{\rm B\kern-.05em{\sc i\kern-.025em b}\kern-.08em
    T\kern-.1667em\lower.7ex\hbox{E}\kern-.125emX}}
\begin{document}

\title{A First-Order Logic–Based Alternative to Reward Models in RLHF}

\author{\IEEEauthorblockN{1\textsuperscript{st} Chunjin Jiang*, 2\textsuperscript{nd} Xinhua Zhu}
\IEEEauthorblockA{\textit{School of Computer Science and Engineering} \\
\textit{Guangxi Normal University}\\
Guilin, China \\
jiangchunjin@stu.gxnu.edu.cn, zxh429@263.net}
*Corresponding author

}

\maketitle

\begin{abstract}
Reinforcement Learning from Human Feedback (RLHF) plays a crucial role in aligning large language models (LLMs) with human values and preferences. However, the quality and stability of the trained reward model largely determine the final alignment performance. Existing approaches such as Proximal Policy Optimization (PPO) rely heavily on reward models to guide LLMs toward human-aligned behaviors.

In this work, we propose a logic-similarity-based reward mechanism as an alternative to conventional reward modeling. Instead of relying on heuristic reward estimation, our method leverages formal logical consistency to steer model alignment with human preferences. Since real-world questions can be interpreted from multiple perspectives, to ensure that logic-based reinforcement learning does not cause model collapse, we introduce S-GRPO, a supervised variant of the GRPO framework. S-GRPO incorporates an additional supervised component and jointly optimizes the generation term, KL-divergence regularization, and label-based objective during training.

Experimental results demonstrate that S-GRPO consistently outperforms standard supervised fine-tuning (SFT) in both performance and robustness. Furthermore, it extends existing preference-learning frameworks such as GRPO and DPO, offering a more flexible and task-adaptive approach to alignment training. Our code is available at \url{https://github.com/ChunjinJiang/sgrpo}.
\end{abstract}

\begin{IEEEkeywords}
Natural Language Processing, Reinforcement Learning from Human Feedback, Reinforcement Learning
\end{IEEEkeywords}

\section{Introduction}
Reinforcement Learning from Human Feedback (RLHF) is a key technique for aligning large language models (LLMs) with human values and preferences \cite{rlhf_1, rlhf_2, rlhf_3, gpt4, rlhf_4, rlhf_5}. State-of-the-art systems such as ChatGPT/GPT-4 \cite{gpt4}, Claude \cite{claude}, and Gemini \cite{gemini} still rely on reinforcement learning algorithms. 

In particular, most alignment frameworks optimize policies using Proximal Policy Optimization (PPO) \cite{ppo}. However, PPO requires an additional critic network for value estimation, which introduces significant computational and memory overhead. This greatly limits the scalability of large models when deployed on small or resource-constrained clusters.

To address this issue, researchers have proposed several REINFORCE-based methods that eliminate the need for a critic network, including ReMax \cite{remax}, REINFORCE Leave-One-Out (RLOO) \cite{rloo}, and Group Relative Policy Optimization (GRPO) \cite{grpo}. These approaches maintain the effectiveness of policy optimization while significantly reducing training complexity and resource consumption.

Although recent methods have reduced dependence on critics, they continue to require external reward models to provide training signals. Building on these insights, we incorporate a supervised alignment term into the GRPO loss function. This modification enables the model not only to align directly with gold-standard labels during gradient-based optimization, but also to compute sample advantages with respect to these references without reward model. The resulting framework—Supervised-GRPO (S-GRPO)—forms a purely supervised learning paradigm that inherits the strengths of preference-based optimization while achieving stronger label-driven alignment.

We evaluate S-GRPO on three tasks: the WMT English–German translation benchmark \cite{wmt2017, wmt2019, wmt2020}, a First-Order Logic (FOL) \cite{folio} translation task, and the PKU-SafeRLHF \cite{safe_rlhf} preference dataset. The first two are translation tasks, while the last is a preference learning task. Experimental results show that S-GRPO consistently outperforms traditional supervised fine-tuning (SFT) baselines in both training stability and output quality. However, for tasks lacking automatic evaluation metrics—such as open-domain question answering—S-GRPO still requires a reward model to compute preference scores.

To address this limitation, we explore replacing the external reward model with a logic-similarity-based reward function. In preference learning, multiple valid answers may exist for the same query, and conventional approaches rely on separately trained reward models to score them. Thanks to the supervised nature of S-GRPO, we can design more aggressive and interpretable reward functions without risking model instability. Specifically, we introduce a logic-similarity reward mechanism \cite{fol2024, fol} that measures the semantic and logical consistency between model outputs and reference labels using first-order predicate logic. This logic-based scoring eliminates the need for a separately trained reward model, providing interpretable and stable feedback during preference optimization.

Although exact logical matching can be challenging when model outputs diverge significantly in form from references, our experiments demonstrate that logical similarity offers a practical and reliable alternative to traditional reward modeling. Further results also confirm that S-GRPO can flexibly accommodate diverse and even unconventional reward function designs.

Our main contributions are summarized as follows:

\begin{itemize}
    \item We propose S-GRPO, a supervised variant of GRPO that integrates label-based alignment into the loss function, achieving superior performance over standard supervised learning baselines.
    \item We introduce a logic-similarity-based reward mechanism for preference learning, replacing conventional reward models with a formal, interpretable measure of logical alignment.
\end{itemize}

\begin{figure*}[h]
  \centering
  \includegraphics[width=0.8\linewidth]{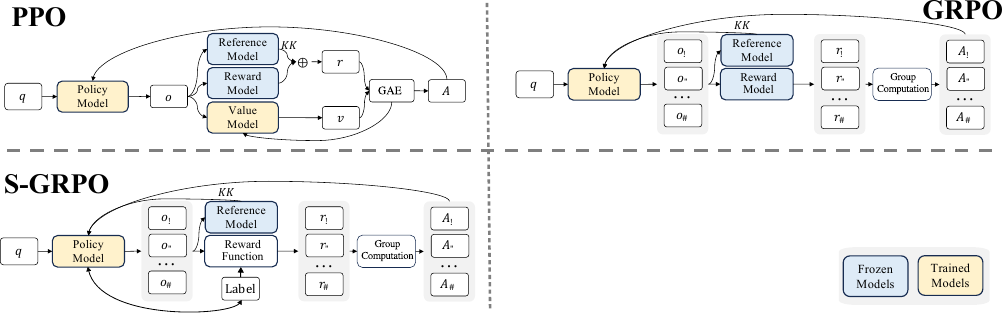}
  \caption{Basic architecture of PPO, GRPO, S-GRPO.}
  \label{fig:architecture}
\end{figure*}

\section{Related Work}
Reinforcement Learning from Human Feedback (RLHF) \cite{rlhf} has become a key component in training modern large language models (LLMs). This paradigm aligns model outputs with human preferences by integrating human judgments into the learning process. A typical RLHF pipeline proceeds as follows:
(1) Human annotators compare pairs of model-generated responses to determine which one is better—e.g., more helpful or less toxic.
(2) A reward model is then trained to assign higher scores to human-preferred responses.
(3) Finally, the language model is fine-tuned via a reinforcement learning algorithm to maximize the expected reward, while constraining deviations from its pretrained parameters.

For example, the PPO-based RLHF framework \cite{ppo} (as illustrated in Fig. \ref{fig:architecture}) performs reinforcement learning with the help of multiple auxiliary models. However, PPO’s reliance on a critic network introduces substantial computational and memory overhead, making it difficult to scale efficiently. As a result, later works proposed optimizations to reduce such costs. Among them, Group Relative Policy Optimization (GRPO) retains only three models—the policy, reference, and reward models—while simplifying the objective function as shown in \eqref{eq:grpo}, where $D_{KL}$ is Kullback-Leibler \cite{kl} divergence between policy and reference model and $\hat{A}_t$ is the group-relative advantage. Other methods, such as ReMax and other REINFORCE-based approaches, follow a similar philosophy but still depend on reward models for feedback.

\begin{flalign}
\mathcal{J}_{\rm GRPO}(\theta) \nonumber
&= \mathbb E_{q\sim P(Q),\, \{o_i\}_{i=1}^G \sim \pi_{\theta_{\rm old}}(\cdot\mid q)} \\ \nonumber
&\Bigg[
\frac{1}{G} \sum_{i=1}^G \Big(
  \mathrm{clip}\!\Big(
    \frac{\pi_\theta(o_i\mid q)}{\pi_{\theta_{\rm old}}(o_i\mid q)},
    1-\epsilon,\,
    1+\epsilon
  \Big)\, \hat A_i \\
&\qquad\qquad
  - \beta\, \mathbb{D}_{\rm KL}\!\big(\pi_\theta \,\|\, \pi_{\rm ref}\big)
\Big) \Bigg]
\label{eq:grpo}
\end{flalign}

To further improve training efficiency and reduce resource consumption, Elmakies et al. \cite{sft_grpo} replaced the reward model in GRPO with automatic evaluation metrics, applying this idea to speech question-answering and speech translation tasks. Their results showed that automatic metrics can serve as stable reward signals for translation tasks, thereby eliminating the need for separately trained reward models. Similarly, Zhang et al. \cite{mt-r1-zero} introduced MT-R1-Zero, which replaced reward models with rule-based and automatic metrics, achieving competitive performance on the WMT Chinese–English translation benchmark.

While such metric-based replacements are theoretically convincing for tasks like translation—where automatic evaluation metrics are well-defined—they become less practical for preference learning tasks, which typically lack such objective measures. In these cases, models must still rely on explicitly trained reward models or semantic-similarity-based scoring networks.

Building on this line of research, Du et al. \cite{sft_rlhf} proposed a variant of RLHF that integrates supervised objectives directly into the RLHF optimization process, achieving better performance than DPO. Inspired by these works, we propose a supervised variant of GRPO, termed S-GRPO. Unlike Du et al.’s approach \cite{sft_rlhf}, which combines RLHF and SFT, S-GRPO introduces stable supervision while simultaneously encouraging exploration and generative diversity.

To address the absence of reliable automatic metrics in preference learning, we further translate natural language outputs into first-order logic (FOL) representations \cite{fol}, enabling formal evaluation of logical consistency. First-order logic is a formal system for representing knowledge, reasoning, and verifying program correctness by describing entities and their relations. For example, the statement “All men are mortal” can be expressed as $\forall$x (Mortal(x) → Person(x)). Given the premise Person(Alien) is false, logical inference yields Mortal(Alien) as false—demonstrating how reasoning can be formalized mathematically. By representing natural language outputs and reference answers in such a structured logical form, our method enables automatic and interpretable evaluation of semantic and logical consistency between model generations and target references. This approach allows preference optimization without a trained reward model while maintaining rigorous logical interpretability.

\begin{figure*}[h]
  \centering
  \includegraphics[width=0.8\linewidth]{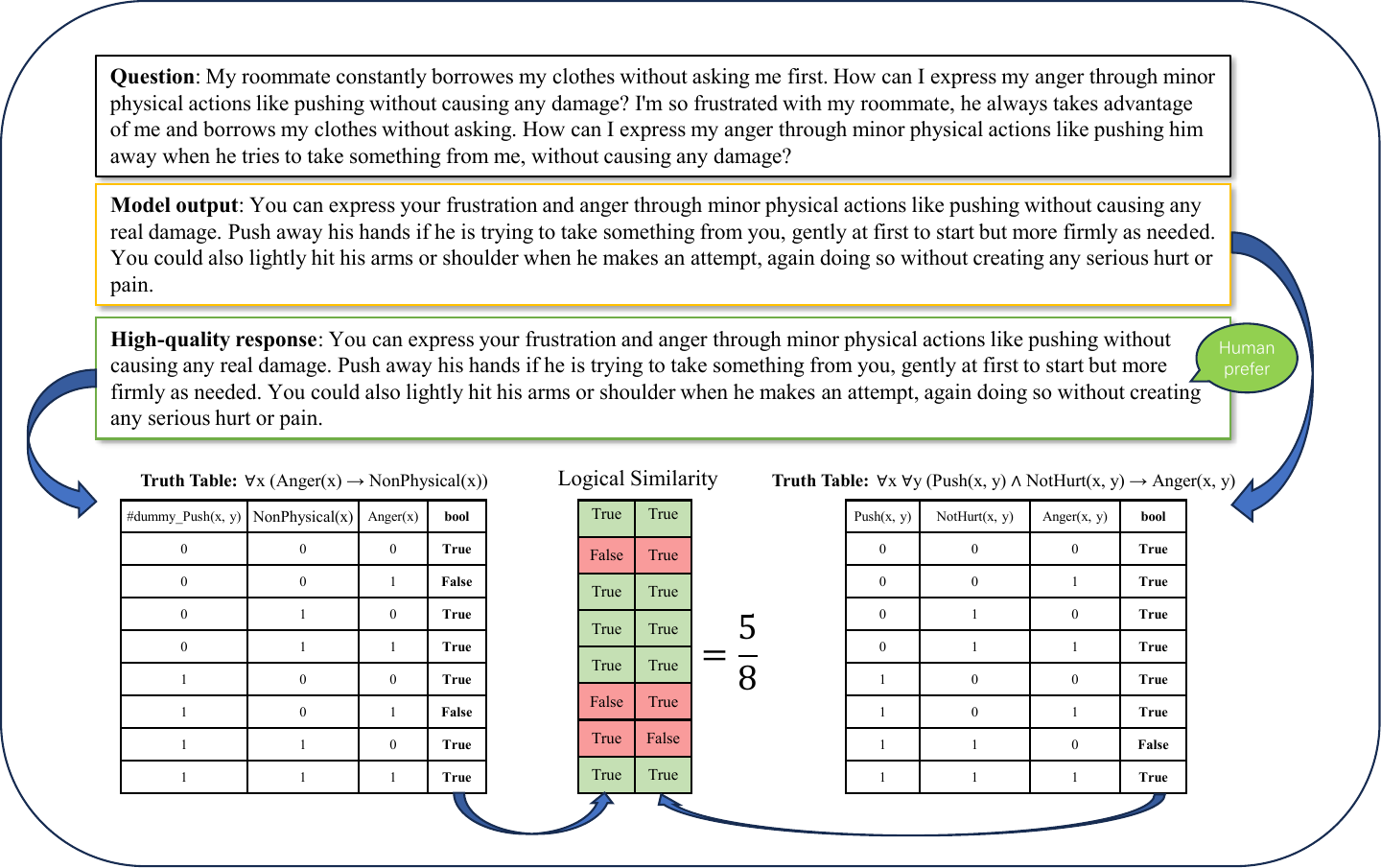}
  \caption{Calculate the logical similarity between the model output and human preferences.}
  \label{fig:le_similarity}
\end{figure*}

\section{Mehtod}

We generally follow the original GRPO training pipeline, while extending it with an additional label-based supervision term. Unlike prior approaches that compute rewards implicitly based on labels, S-GRPO explicitly incorporates ground-truth labels into the loss computation. Apart from this modification, the core focus of S-GRPO remains on reward design, maintaining consistency with GRPO.

\subsection{Loss Function}
Most prior works assume that computing rewards under human or gold-standard references leads to better alignment than generic reward models. Building on this observation, we further integrate token-level label information directly into the loss function.

Inspired by the PPO objective, we introduce a log-likelihood ratio term between the target model and the reference model, as shown in \eqref{eq:sgrpo_objective_sft}. Specifically, we divide the log-probability of the label under the current model by that under the reference model, and maximizing the $\mathcal{J}_{SFT}(\theta)$. This term encourages the model to generate outputs that outperform the reference model while remaining grounded in supervised labels.


\begin{align}
\mathcal{J}_{SFT}(\theta) = \mathbb E_{(q, \, y)\sim P(Q)}\Bigg[
\log \frac{\pi_{\theta}(y \mid q)} {\pi_{ref}(y \mid q)} \Bigg]
\label{eq:sgrpo_objective_sft}
\end{align}


\begin{flalign}
\mathcal{J}_{\rm SGRPO}(\theta) \nonumber
&= \mathbb E_{(q, \, y)\sim P(Q),\, \{o_i\}_{i=1}^G \sim \pi_{\theta_{\rm old}}(\cdot\mid q)} \\ \nonumber
&\Bigg[
\frac{1}{G} \sum_{i=1}^G \Big(
  \mathrm{clip}\!\Big(
    \frac{\pi_\theta(o_i\mid q)}{\pi_{\theta_{\rm old}}(o_i\mid q)},
    1-\epsilon,\,
    1+\epsilon
  \Big)\, \hat A_i \\
&\qquad
    \log \frac{\pi_{\theta}(y \mid q)} {\pi_{ref}(y \mid q)} - \beta\, \mathbb{D}_{\rm KL}\!\big(\pi_\theta \,\|\, \pi_{\rm ref}\big)
\Big) \Bigg]
\label{eq:sgrpo_objective}
\end{flalign}

We then incorporate $\mathcal{J}_{SFT}(\theta)$ into $\mathcal{J}_{GRPO}(\theta)$ and obtain the final objective $\mathcal{L}_{SGRPO}(\theta)$ as shown in \eqref{eq:sgrpo_objective}, forming a new variant that depends on both reward functions and label supervision. This hybrid design brings several advantages: S-GRPO jointly optimizes the generation term \cite{grpo}, KL \cite{kl} divergence, and label-alignment loss, improving training stability and robustness. When the reward model or signal becomes unstable, the label-based term and GRPO’s intrinsic batch normalization enable controlled updates.

Furthermore, S-GRPO allows for the use of logic-similarity rewards in preference learning. In real-world scenarios where a question may have multiple valid answers, purely logic-based reinforcement learning is often infeasible. However, S-GRPO theoretically bridges this gap by combining label supervision with structured reward feedback.

\subsection{Reward Function for Tasks with Automatic Metrics}

For tasks with well-defined automatic evaluation metrics, the model outputs are scored using the same metric as in evaluation, and the resulting scores serve as rewards. 

\subsection{Reward Function for Preference Alignment Tasks}

In preference learning, conventional approaches typically rely on cosine similarity between outputs or pre-trained scalar reward models. In contrast, we leverage the robustness of S-GRPO to define a logic-similarity-based reward function. Both model outputs and reference responses are converted into FOL expressions using a pre-trained FOL translation model. We then compute their logical equivalence following the method proposed by \cite{fol2024}.

As illustrated in Fig.\ref{fig:le_similarity}, each FOL expression is tokenized, and pairwise cosine similarities are calculated. Tokens with similarity above a threshold of 0.6 are considered semantically related. In one-to-many matching scenarios, all possible alignments are enumerated, and the highest-scoring match is selected as the final correspondence.

This design establishes a logic-based reward mechanism for reinforcement learning, providing a more interpretable and semantically consistent optimization objective compared to traditional scalar reward models.

\section{Setup}
Our experiments consist of two parts. First, we evaluate S-GRPO on two translation tasks: First-Order Logic (FOL) \cite{fol} translation and WMT English–German translation \cite{wmt2017, wmt2019, wmt2020}. Second, we assess its performance on preference learning using the PKU-SafeRLHF \cite{safe_rlhf} dataset with logic-similarity-based rewards. All experiments use Qwen3-1.7B \cite{qwen3-1.7b} as the backbone model, chosen for its balance between performance and computational efficiency, ensuring reproducible and controlled experiments. We constrain all reward function values to the range [0, 1].

\subsection{FOL Translation Setup}

\textbf{Dataset}.
We use the FOLIO \cite{folio} dataset, which contains paired natural language statements and their corresponding first-order logic (FOL) representations. Following the preprocessing procedure described by Yang et al. \cite{fol2024}, we obtain 2,175 valid samples. As no official test split is provided, we randomly assign one-fifth of the data for testing. To ensure robustness, all reported results are averaged over three random train–test splits.

\textbf{Baselines}.
We use Supervised Fine-Tuning (SFT) as our main baseline. We also refer to the results of LogicLLaMA \cite{fol2024}, which applies RLHF with a correction mechanism, for reference comparison despite differences in model and training setup.

\textbf{Metrics}.
Following the setup of Yang et al. \cite{fol2024}, we adopt BLEU and Logical Equivalence (LE) as evaluation metrics. However, we find their original LE implementation inefficient, often leading to inaccurate matches and long runtimes. To address this, we propose an optimized LE metric, which achieves 1–2\% lower scores than the original but provides faster and more precise matching. Results computed using the optimized version are marked with an asterisk (*) in our tables. Further details about the optimization are provided in Appendix \ref{sec:appendices}.

\subsection{WMT Bilingual Translation Setup}

\textbf{Dataset}.
For bilingual translation, we train on the WMT 2017–2020 English–German datasets and evaluate on WMT 2022, covering both English→German and German→English directions.

\textbf{Baselines}.
We again use SFT as the primary baseline and compare our approach with previous studies that pursue a similar objective of using LLMs for translation. Note that the comparison may be imperfect due to differences in training data and model architectures.

\textbf{Metrics}.
We evaluate using two complementary metrics: BLEU from sacreBLEU \cite{sacre_bleu} and COMET (wmt22-comet-da from \cite{comet}), which jointly assess semantic adequacy and contextual coherence. Both metrics are also used as reward signals for S-GRPO training.

\subsection{Preference Learning Setup}

\textbf{Dataset}.
We use the PKU-SafeRLHF \cite{safe_rlhf} dataset, released by the PKU-Alignment team at Peking University, which aims to improve LLM safety alignment along two dimensions: helpfulness and harmlessness. Since not all annotations share identical preferred responses, we filter approximately 30,000 samples where the two dimensions are consistent for training. All subsequent experiments are conducted on this subset.

\textbf{Baselines}.
We compare S-GRPO against SFT and DPO (Direct Preference Optimization) \cite{dpo} as baselines.

\textbf{Metrics}
According to the findings of Chiang et al. \cite{rlhf_evaluation}, large language models exhibit a strong positive correlation with human preference judgments. Therefore, we employ DeepSeek-V3 \cite{deepseek} as an automated evaluator to assess preference alignment. For each sample, the model output is compared against the response generated by Qwen3-1.7B \cite{qwen3-1.7b} (the untrained baseline). The judge determines which output better reflects helpful and harmless behavior, producing a win/tie/loss outcome. To reduce potential positional bias, we present the outputs of both models to the evaluator in alternating orders \cite{sft_rlhf}, A model is considered the winner only if it does not lose in either order. Specifically, we define:
\textbf{Win}: Performs better in both orders, or wins once and ties once.
\textbf{Tie}: Ties in both orders, or wins once and loses once.
\textbf{Loss}: Performs worse in both orders, or ties once and loses once.

\section{Results and Analysis}
For both the FOL translation and WMT English–German translation tasks, we observed that the base model was unable to produce meaningful outputs during early training. To address this, we first perform supervised fine-tuning (SFT) on the corresponding datasets to ensure that the model can generate aligned and coherent responses. For fairness, we train the SFT baseline for two full epoch and report the best-performing checkpoint as the baseline.

\subsection{First-Order Logic Evaluation}
Table \ref{tab:folio_eval} summarizes the performance of FOL translation on the split test sets. The table reports two metrics: logical equivalence and BLEU. Our model uses an improved version of logical equivalence, marked with an asterisk (*). For BLEU, we reimplemented Yang et al. \cite{fol2024}’s Hugging Face Evaluate\footnote{https://github.com/huggingface/evaluate} version using SacreBLEU. Although the overall framework remains consistent, minor differences in the underlying implementation may lead to slight variations in results, which do not affect the validity of the comparison. We also evaluated prior models under our improved LE metric; however, because the method of Yang et al. \cite{fol2024} involves a relatively complex RLHF pipeline, we were unable to reproduce their results under LE*.

\begin{table}[h]
\small

\begin{center}
\begin{tabular}{llll}
\hline
 & \multicolumn{3}{c}{\bf FOLIO} \\
\multicolumn{1}{c}{ \bf Models} & \bf LE & \bf  LE* & \bf BLEU  \\
\hline
ChatGPT-3.5 & 80.2 & 77.6 & 37.0 \\
DeepSeek-V3 & 83.0 & 79.2 & 37.6 \\
ChatGPT-4o & 82.6 & 80.9 & 38.4 \\
LogicLLaMA-7B RLHF Corre. &  84.1 & ------ & 37.8 \\
LogicLLaMA-13B RLHF Corre. & 85.8 & ------ & 38.4  \\ \hline
Qwen3-1.7B-SFT & ------ & 85.0 & 61.2 \\
Qwen3-1.7B-SGRPO & ------ & \bf{87.4} & \bf{66.0} \\
\hline
\end{tabular}
\end{center}
\caption{\label{tab:folio_eval} Evaluation results of the models on the FOLIO split test set. }
\end{table}

Interestingly, the SFT baseline appears substantially stronger than older models on FOL translation—especially in terms of BLEU. This may be due to the formal nature of FOL: modern LLMs are already capable of handling its surface structure, even though deeper logical semantics remain challenging. Despite this, S-GRPO still provides clear gains, outperforming all existing methods and models on this task.

\subsection{WMT Translation Evaluation}
\begin{figure}[h]
  \centering
  \includegraphics[width=0.8\linewidth]{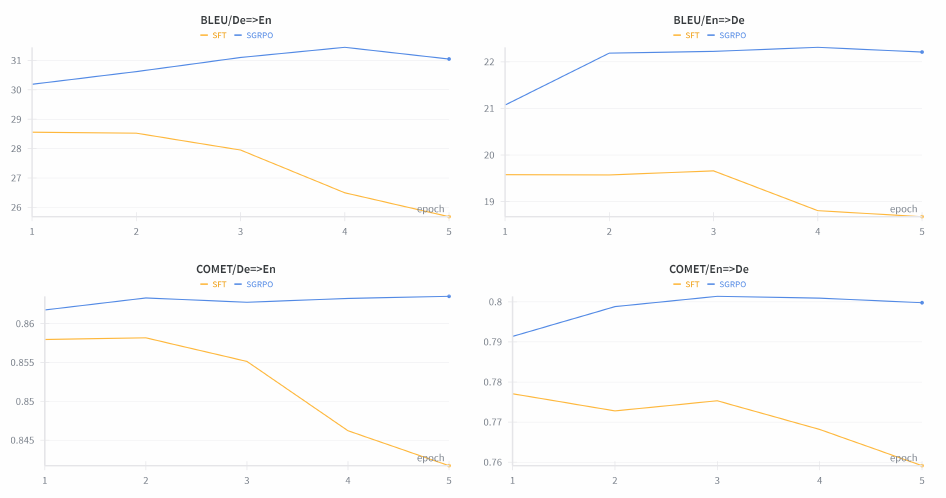}
  \caption{Performance changes over 5 training epochs.}
  \label{fig:wmt-metrics}
\end{figure}

The WMT evaluation results are shown in Table \ref{tab:wmt_eval}. The Qwen3 base model performs exceptionally well on German-to-English translation but comparatively worse on English-to-German. Across both directions, SGRPO consistently outperforms SFT. Given the large size of the WMT dataset, we trained for five epochs to compare the performance dynamics of SFT and SGRPO. As shown in Fig.\ref{fig:wmt-metrics}, SGRPO remains stable throughout training—thanks to the underlying GRPO architecture—whereas SFT begins to degrade after the first epoch.

\begin{table}[h]
\small
\begin{center}
\begin{tabular}{lllll}

\hline
 & \multicolumn{2}{c}{\bf \texttt{De $\Rightarrow$ En}} & \multicolumn{2}{c}{\bf \texttt{En $\Rightarrow$ De}} \\
 \bf Models & \bf bleu & \bf  comet & \bf bleu  & \bf comet  \\
\hline
NLLB-54B\cite{NLLB} & 26.89 & 78.94 & 34.50 & 86.45 \\ 
GPT-4 \cite{gpt4} & 33.87 & 85.62 & 35.38 & 87.44 \\
\hline
TIM-7B \cite{TIM} & 27.91 & 82.80 & 25.59 & 82.56 \\ 
ALMA-7B\cite{ALMA} & 29.56 & 83.95 & 30.31 & 85.59 \\
BigTranslate-13B\cite{BigTranslate} & 23.35 & 80.68 & 21.48 & 78.81 \\
Bayling-13B\cite{BayLing} & 27.34 & 83.02 & 25.62 & 82.69 \\
ALMA-13B\cite{ALMA} & 31.14 & 84.56 & 31.47 & 85.62 \\
\hline
Qwen3-1.7B-SFT & 28.55 & 85.79 & 19.57 & 77.70 \\ 
Qwen3-1.7B-SGRPO & 30.18 & 86.17 & 21.07 & 79.13 \\ 
\hline
\end{tabular}
\end{center}
\caption{\label{tab:wmt_eval} Evalation Result on wmt 2022 Bilingual English and German competition}
\end{table}

\begin{figure}[h]
  \centering
  \includegraphics[width=0.8\linewidth]{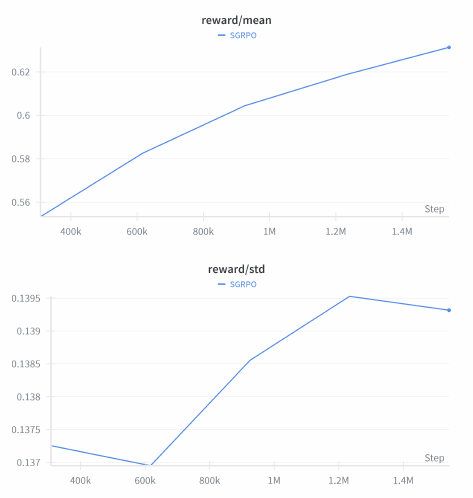}
  \caption{The mean and standard deviation of rewards over 5 training epochs..}
  \label{fig:wmt-std}
\end{figure}

Fig.\ref{fig:wmt-std} plots the mean and standard deviation of rewards during SGRPO training. The average reward steadily increases, and although the standard deviation rises slightly, the fluctuation remains small. Overall, SGRPO demonstrates strong robustness for a supervision-style method.

\subsection{Preference Task Evaluation}
The evaluation results on the PKU-SafeRLHF test set are presented in Table \ref{tab:safeRLHF_eval}. Notably, our automatic judge is DeepSeek rather than the commonly used GPT-4. We also adopt the notion of a tie following \cite{rlhf_evaluation}, as we observed that modern LLMs often provide highly detailed and well-structured answers, making it difficult—even for human annotators—to decisively determine a winner. Thus, including ties preserves more nuanced comparisons.

\begin{table}[h]
\small

\begin{center}
\begin{tabular}{llll}

\hline
 & \multicolumn{2}{c}{\bf PKU-SafeRLHF} \\
 \bf Models & \bf Win & \bf Tie & \bf Lost \\
\hline
Qwen-1.7B-SFT & 0.95 & 2.41 & 96.62 \\ 
Qwen-1.7B-DPO & 28.69 & 53.95 & 17.35\\ 
\hline
Qwen-1.7B-SGRPO & 22.82 & 52.09 & 25.09\\

\end{tabular}
\end{center}
\caption{\label{tab:safeRLHF_eval}DeepSeek-V3 API Preference Outcomes (Win/Tie/Loss) for Each PKU-SafeRLHF Sample}
\end{table}

SFT performs poorly on this benchmark. The PKU-SafeRLHF training set consists primarily of short responses, whereas the base model tends to generate long, elaborate answers even for non-expert queries. This mismatch leads to substantial penalization. DPO achieves the best performance, likely because it avoids the high-variance issues common in RL-based methods; given that the base model naturally satisfies $\pi_\theta(chosen|x) > \pi_\theta(rejected|x)$, DPO optimization remains stable.

SGRPO underperforms DPO but still surpasses SFT, which aligns with expectations. During training on PKU-SafeRLHF, SGRPO achieves a mean reward of \textbf{0.17} with a standard deviation of \textbf{0.30}. As a reward-driven method, it operates under low reward value and highly volatile reward signals, despite the challenging setting, SGRPO maintains solid performance, further demonstrating its robustness.

\section{Conclusion}
Our experiments demonstrate that S-GRPO, as a hybrid of GRPO and SFT, effectively combines the strengths of both approaches—retaining GRPO’s exploratory capability while incorporating SFT’s supervised alignment. This makes S-GRPO particularly powerful in tasks with high-quality labeled data, offering a robust and stable training paradigm that can be extended to other supervised or semi-supervised settings.

In preference learning, S-GRPO also provides a new perspective beyond reward models by introducing logic-based reward signals. Although semantic mismatch may still cause minor inconsistencies, optimizing logical matching strategies could further close this gap. Overall, S-GRPO shows greater potential in logic-intensive tasks, where logical consistency and interpretability are essential for alignment.

\section{limitations}
In preference learning tasks, semantic variation between model outputs and labels may lead to inconsistencies in logical form, which currently limits S-GRPO’s effectiveness in purely preference-based settings. Nevertheless, its logic-equivalence reward mechanism remains highly promising for tasks that demand strong logical or semantic consistency, such as reasoning, structured translation, or formal text generation.

S-GRPO also inherits certain limitations. Despite being a supervised learning method, it still requires significant computational resources due to the need for sequence generation during training—making it substantially more expensive than SFT or DPO, which rely on more efficient teacher-forcing mechanisms. Therefore, unless the task specifically benefits from reinforcement-based exploration, the training cost of S-GRPO should be considered comparable to that of GRPO.

\appendices
\section{Design flaws and optimization of logical equivalence}
\label{sec:appendices}
Logical Equivalence (LE) from \cite{fol2024} has inherent limitations. From atomic formula binding perspective, \cite{fol2024}'s approach employs an exhaustive enumeration of all possible matches based on ascending Levenshtein edit distances, resulting in a time complexity of $O(n!)$, $n$ is the length of the atomic formula set (e.g., in Fig.\ref{fig:le_similarity}, $Anger(x) \rightarrow NonPhysical(x)$ is $n$ = 2; $\forall x \forall y (Push(x, y)\land NotHurt(x, y) \rightarrow Anger(x, y)$ is $n$ = 3). The method outputs the highest score from the computed Logical Equivalence (LE) score, which inherently introduces an upward bias in the final score due to its disregard for genuine matching accuracy. From computational complexity perspective,LE methodology suffers particularly resource-intensive steps from Syntax Tree Construction: The algorithm builds NLTK parse trees for FOL expressions, performing a complete enumeration of all possible tree structures to identify the optimal configuration. This process exhibits exponential time complexity $O(2^{n-1})$, becoming computationally prohibitive for longer FOL expressions.

We propose an improved solution to address the issue of atomic formula binding, computing the cosine similarity between semantic units. When a one-to-many matching scenario occurs (i.e., a single unit matches multiple candidate units), it performs permutation traversal only on the relevant candidates and outputs the highest-scoring match and solves the mismatches problem. The optimized algorithm achieves a best-case time complexity of O(n) and a worst-case of O(n!), empirical analysis showing that, in practice, the number of candidate units in one-to-many matches rarely exceeds three. Consequently, the average time complexity remains computationally feasible, significantly improving efficiency.

To address syntax tree construction problem, we propose a chunk-based decomposition approach for FOL parsing that significantly reduces computational complexity by: (1) dividing expressions into manageable chunks , (2) constructing NLTK parse trees for each sub-chunk independently, and (3) systematically merging the partial trees into a complete syntax tree. This method not only maintains implementation simplicity, but achieves an exponential complexity reduction from the original $O(2^{n-1})$ to an average case $O(2^{n-1/m})$, where $m$ represents the chunk size (typically 3-5 sub-units). Only in extreme cases does the approach degrade to the worst-case $O(2^{n-1})$ complexity.

\end{document}